\def\year{2019}\relax
\documentclass[letterpaper]{article} %DO NOT CHANGE THIS
\usepackage{aaai19}  %Required
\usepackage{times}  %Required
\usepackage{helvet}  %Required
\usepackage{courier}  %Required
\usepackage{url}  %Required
\usepackage{graphicx}  %Required
\usepackage[T1]{fontenc} %Added
\usepackage{algorithm} %Added
\usepackage{algpseudocode} %Added
\usepackage{booktabs} %Added
\usepackage{subcaption}
\usepackage{times}
\usepackage{epsfig}
\usepackage{threeparttable}
\usepackage{enumitem}
\usepackage{xcolor}
\usepackage{subcaption}

\usepackage[font={small}]{caption}
\usepackage{hyperref}

\begin{document}
\nocopyright
% The file aaai.sty is the style file for AAAI Press
% proceedings, working notes, and technical reports.
%
\title{Adversarial Framing for Image and Video Classification}
\author{Konrad \.Zo\l{}na$^*$\textsuperscript{, 1, 2},
Micha\l{} Zaj\k{a}c\thanks{Equal contribution}\textsuperscript{, 1, 3},
Negar Rostamzadeh\textsuperscript{2},
Pedro O. Pinheiro\textsuperscript{2}\\
\textsuperscript{1}{Jagiellonian University, Krak\'ow, Poland}\\
\textsuperscript{2}{Element AI, Montr\'eal, Canada}\\
\textsuperscript{3}{Nomagic, Warsaw, Poland}\\
konrad.zolna@gmail.com,
emzajac@gmail.com,
negar@elementai.com,
pedro@opinheiro.com
}
\maketitle

\begin{abstract}
Neural networks are prone to adversarial attacks. In general, such attacks deteriorate the quality of the input by either slightly modifying most of its pixels, or by occluding it with a patch. In this paper, we propose a method that keeps the image unchanged and only adds an \emph{adversarial framing} on the border of the image. We show empirically that our method is able to successfully attack state-of-the-art methods on both image and video classification problems. Notably, the proposed method results in a universal attack which is very fast at test time. Source code can be found at \href{https://github.com/zajaczajac/adv\_framing}{\texttt{github.com/zajaczajac/adv\_framing}}.
\end{abstract}

%%%%%%%%%%%%%%%%%%%%%%%%%%%%%%%%%%%%%%%%%%%%%%%%%%%%%%%%%%%%%%%%%%%%%%%%%%%%%%%%%%%%%%%%%%%%%%%%%%
% Introduction
%%%%%%%%%%%%%%%%%%%%%%%%%%%%%%%%%%%%%%%%%%%%%%%%%%%%%%%%%%%%%%%%%%%%%%%%%%%%%%%%%%%%%%%%%%%%%%%%%%
\section{Introduction}

The remarkable success of deep convolutional networks for image and video classification \cite{karpathy2014large,krizhevsky2012imagenet} has spurred interest in analyzing their robustness. Unfortunately, it turned out that even though neural networks often achieve human level performance \cite{taigman2014deepface}, they are susceptible to adversarial attacks~\cite{original}. It means that the output of a neural network-based classifier may be drastically changed by applying a small perturbation to its input. We divide such perturbations into two categories: \emph{fully-affecting} and \emph{partially-affecting}.
\begin{itemize}
    \item Fully-affecting attacks generate small pixel intensity modifications which are optimized to be hardly visible for humans. These attacks typically have their $\ell_2$ or $\ell_\infty$ norm constrained~\cite{towards_evaluating,deepfool} and hence affect the whole image.
    \item Partially-affecting attacks usually have their $\ell_0$ norm constrained. They introduce perceptible but small occlusion to the image, such as a patch~\cite{adversarial_patch,lavan} or a single pixel~\cite{single_pixel}.
\end{itemize}

The attacks mentioned above either slightly modify all the pixels of the image or occlude parts of it. However, the attackers may find this to be a serious limitation and seek for new types of attacks. For instance, consider a scenario where they upload videos containing forbidden content. Their goal is to bypass video-sharing website's filters. At the same time, the perturbations introduced should not be distracting and all information should be retained.

\begin{figure*}[t]
\centering
\includegraphics[width=\linewidth]{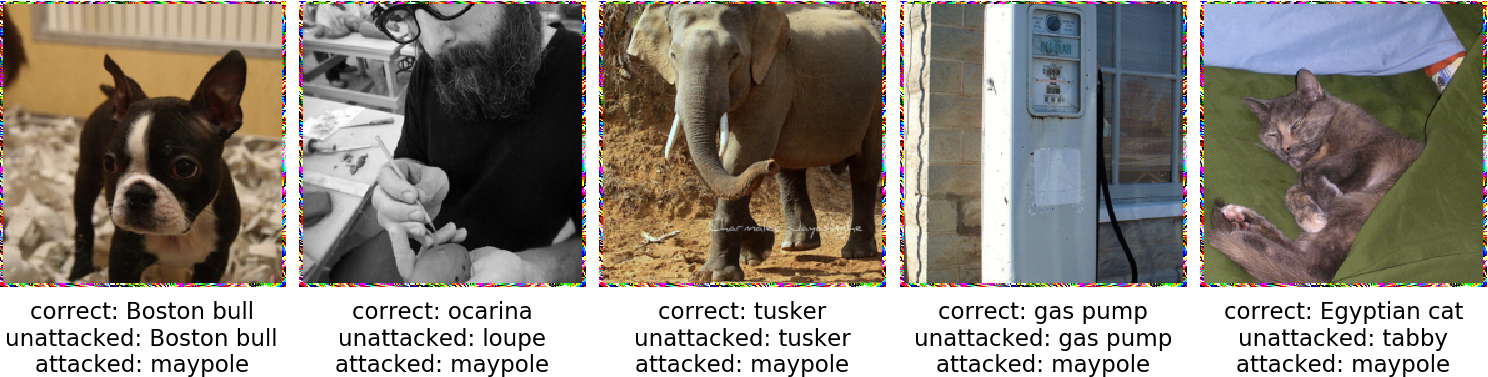}
\caption{Examples from ImageNet with adversarial framing of width $3$. Most of the images are wrongly classified as a maypole. We hypothesize that the colorfulness of that class makes it especially easy for \textbf{AF} to resemble it.
\label{fig:imagenet_examples}
}
\end{figure*}

In this paper, a new attack which is well-suited for the above-mentioned purposes is demonstrated. The method, dubbed adversarial framing (\textbf{AF}), consists in simply adding a thin border around the original input (which may be an image or a video), keeping the whole content unchanged (see Figure~\ref{fig:imagenet_examples} and \href{https://youtu.be/PrU9R6eFNTs}{\texttt{youtu.be/PrU9R6eFNTs}} for some qualitative results). The attack is universal~\cite{universal}, which means the same \textbf{AF} is applied to all inputs. The method only requires substantial computing during the training procedure. At test time, the only extra computation required is the appending of the precomputed framing to the input.

Similarly to \cite{athalye2018obfuscated}, we believe that research on attack techniques deepens understanding of inner workings of neural networks. We hope that our work and analyzing adversarial attacks in general can be helpful in designing defenses and/or robust methods.

In this work, we consider a white-box setting, in which an access to the architecture and weights of the trained classifier is given. Previous work has shown that if only black-box access is given, a surrogate model can be leveraged to obtain an attack that transfers well to the original model~\cite{blackbox}. Therefore, a white-box model is a realistic assumption and, in fact, is the most commonly considered paradigm in the literature.
%%%%%%%%%%%%%%%%%%%%%%%%%%%%%%%%%%%%%%%%%%%%%%%%%%%%%%%%%%%%%%%%%%%%%%%%%%%%%%%%%%%%%%%%%%%%%%%%%%
% Method
%%%%%%%%%%%%%%%%%%%%%%%%%%%%%%%%%%%%%%%%%%%%%%%%%%%%%%%%%%%%%%%%%%%%%%%%%%%%%%%%%%%%%%%%%%%%%%%%%%
\section{Method}

\subsection{Computing the adversarial framing}

Suppose a labeled dataset of images or videos $\mathcal{D}$ is given. Moreover, a differentiable classifier $f$ has been trained so that for each input $x$ and class $c$, a probability $f_c(x)$ is assigned to $x$ of being in class $c$.

We now present a procedure to train the adversarial framing to attack $f$. During training, a minibatch is sampled from $\mathcal{D}$. Every example is surrounded with the same framing, which is the current version of the trained \textbf{AF}. In case of videos, every frame of each example is surrounded with the same framing. Then the classification loss is backpropagated and the framing is modified using its gradients to maximize the loss. The training continues until convergence. The framing's width $W$ is a tunable hyperparameter fixed at the beginning of the training procedure.

For a detailed explanation see Algorithm~\ref{alg:adversarial_framing}. The algorithm is presented for image datasets. The modification for video datasets is straightforward.

Note that the input size is modified due to the addition of the framing. This does not pose any issue to the CNN-based classifier, as most modern architectures (such as ResNet \cite{resnet} or ResNeXt \cite{resnext}) accept various input sizes.
If the classifier's input size is fixed, the proposed algorithm can be simply modified so that the image is resized before applying adversarial framing. We investigate performance under various resizing strategies further in the paper.

\begin{algorithm}[t]
\caption{Training of the adversarial framing}
\begin{algorithmic}[1]
\State \textbf{input:} Dataset $\mathcal{D} = \{(x_i, y_i)\}$, $x_i \in [0, 1]^{h \times w \times 3}$, classifier $f$, framing's width $W$
\State \textbf{output:} Universal adversarial framing $\theta$
\State Initialize $\hat{\theta} \sim \mathcal{N}(0, 1)$, of size $2W(h + w + 2W)$
\Repeat
\For{each datapoint $(x_i, y_i) \in \mathcal{D}$}
\State $\hat{x}_i \gets x_i$ surrounded by $\theta := Sigmoid(\hat{\theta})$
\EndFor
\State update $\hat{\theta}$ to minimize $\frac{1}{|\mathcal{D}|}\sum_{i} \log(f_{y_i}(\hat{x}_i))$
\Until{convergence}
\end{algorithmic}
\label{alg:adversarial_framing}
\end{algorithm}

%%%%%%%%%%%%%%%%%%%%%%%%%%%%%%%%%%%%%%%%%%%%%%%%%%%%%%%%%%%%%%%%%%%%%%%%%%%%%%%%%%%%%%%%%%%%%%%%%%
% Experiments
%%%%%%%%%%%%%%%%%%%%%%%%%%%%%%%%%%%%%%%%%%%%%%%%%%%%%%%%%%%%%%%%%%%%%%%%%%%%%%%%%%%%%%%%%%%%%%%%%%
\section{Experiments}
\subsection{Untargeted attacks}
We performed untargeted attacks against state-of-the-art classifiers for ImageNet \cite{imagenet} and UCF101 \cite{ucf101} datasets. We compare our \textbf{AF} to two simple baselines. They both do not require any training and are fixed. One applies uniformly distributed random noise (\textbf{RF}) and another black pixels only (\textbf{BF}).

ImageNet is a large-scale image dataset containing over million images from 1000 various classes. It serves as a popular benchmark for image classification. We performed attacks against ResNet-50 \cite{resnet} model pretrained on ImageNet. The model was taken from PyTorch Model Zoo \cite{pytorch}. Results are reported in Table \ref{imagenet_untargeted}.

UCF101 is a dataset containing realistic videos. Each video contains a person performing some action, out of 101 possible classes. We tested our method by performing an attack on a ResNeXt-101 based spatio-temporal 3D CNN -- we used model pretrained by \cite{resnet_3d}. This model takes clips as input, each containing 16 consecutive frames. Results are reported in Table \ref{ucf101_untargeted}.

\begin{table*}[t]
    \begin{subtable}{.5\linewidth}
        \centering
        % \small
        \tabcolsep=0.14cm
        \begin{tabular}{l|cccc}
         \toprule
        \textbf{Attack} & $W = 1$ & $W = 2$ & $W = 3$ & $W = 4$ \\
         \midrule

         \textbf{None} & \multicolumn{4}{c}{76.13\%} \\
         \midrule
         \textbf{RF} & $70.13\%$ & $67.63\%$ & $68.36\%$ & $67.25\%$ \\
         \textbf{BF} & $72.99\%$ & $72.9\%$ & $72.39\%$ & $72.34\%$ \\
         \midrule
         \textbf{AF} & $10.53\%$ & $0.44\%$ & $0.11\%$ & $0.1\%$ \\
         \bottomrule
        \end{tabular}
        \caption{ImageNet dataset}
          \label{imagenet_untargeted}
    \end{subtable}
    \begin{subtable}{.5\linewidth}
        \centering
        % \small
        \tabcolsep=0.14cm
        \begin{tabular}{l|cccc}
         \toprule
        \textbf{Attack} & $W = 1$ & $W = 2$ & $W = 3$ & $W = 4$ \\
         \midrule

         \textbf{None} & \multicolumn{4}{c}{$85.95\%$}  \\
         \midrule
         \textbf{RF} & $82.57\%$ & $80.53\%$ & $81.11\%$ & $79.74\%$ \\
         \textbf{BF} & $84.94\%$ & $84.73\%$ & $84.75\%$ & $84.59\%$ \\
         \midrule
         \textbf{AF} & $65.77\%$ & $22.12\%$ & $9.45\%$ & $2.05\%$ \\
         \bottomrule
        \end{tabular}
        \caption{UCF101 dataset}
          \label{ucf101_untargeted}
    \end{subtable}
    \caption{Accuracies of the classifiers (full validation set)
        for various values of the framing width $W$.}
\end{table*}

\subsection{Targeted attacks}

As it can be seen in Figure~\ref{fig:imagenet_examples} and Figure~\ref{fig_saliency}, adversarial framing usually fools the classifier into wrongly recognizing one particular class. In the case of ImageNet, this adversarial class is usually maypole -- even across different trainings. We hypothesize that this is because of colorfulness of this object.

In order to make sure that the performance of our attack does not depend on presence of such special classes, we performed attacks in targeted setting. In these experiments, instead of minimizing the output score for the ground-truth class, we maximize the score for a randomly selected target class. We report success rate (\emph{i.e.\ }percentage of images classified as a given target) for different target classes.

Results for ImageNet dataset are in Table~\ref{imagenet_targeted} and for UCF101 dataset are in Table~\ref{ucf101_targeted}.

\begin{table*}[t]
    \begin{subtable}{.5\linewidth}
        \centering
        % \small
        \tabcolsep=0.14cm
        \begin{tabular}{l|ccc}
         \toprule
        & \textbf{min} & \textbf{avg} & \textbf{max} \\
         \midrule
         \textbf{AF}, $W = 4$ & $99.15\%$ & $99.66\%$ & $99.98\%$ \\
         \bottomrule
        \end{tabular}
        \vskip 0.1in
        \caption{ImageNet dataset}
          \label{imagenet_targeted}
    \end{subtable}
    \begin{subtable}{.5\linewidth}
        \centering
        % \small
        \tabcolsep=0.14cm
        \begin{tabular}{l|ccc}
         \toprule
         & \textbf{min} & \textbf{avg} & \textbf{max} \\
         \midrule

         \textbf{AF}, $W = 4$ & $73.04\%$ & $89.63\%$ & $99.78\%$ \\
         \bottomrule
        \end{tabular}
        \vskip 0.1in
        \caption{UCF101 dataset}
          \label{ucf101_targeted}
    \end{subtable}
    \caption{Success rate of targeted attacks (the higher the better) with adversarial framing of width 4. Minimum, average and maximum values are taken across 8 different targets.}
\end{table*}

\subsection{Training details}

In all the experiments we used Adam optimizer \cite{adam}. For all the hyperparameters of the optimizer except for learning rate, we used default values from PyTorch \cite{pytorch} implementation.

On ImageNet \cite{imagenet} we trained for 5 epochs, with initial learning rate 0.1 decaying by 0.1 every 2 epochs and batch size 32. On UCF101 \cite{ucf101} we trained for 60 epochs, with initial learning rate 0.03 decaying by 0.3 every 15 epochs and batch size 32. On both these datasets, we trained adversarial framing using training data only. All the reported results were computed on validation data.

On ImageNet, we applied the framing to images previously resized to $224 \times 224$. On UCF101, we applied the framing to images  previously resized to $112 \times 112$. These are standard input dimensions for aforementioned datasets.

\section{Further analyses}

\subsection{Saliency visualization}
Grad-CAM~\cite{selvaraju2017} is a method for producing visual explanations for a convolutional neural network's predictions. For a given classifier $f$, input $x$ and a class $c$, it computes a heatmap visualizing how much particular regions of $x$ contribute to a score of the class $c$ output by $f$.

\begin{figure*}
\renewcommand{\familydefault}{\sfdefault}
\captionsetup[subfigure]{labelformat=empty}
% \captionsetup[subfigure]{position=top}
    \centering
    \begin{subfigure}[b]{0.25\textwidth}
        \centering
        \includegraphics[width=0.97\textwidth]{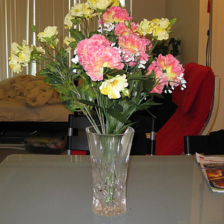}
        \caption{Ground-truth: vase}
    \end{subfigure}\hspace{0.025\textwidth}
    \begin{subfigure}[b]{0.25\textwidth}
        \centering
        \includegraphics[width=0.97\textwidth]{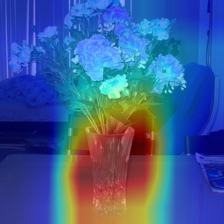}
        \caption{Predicted: vase}
    \end{subfigure}\hspace{0.025\textwidth}
    \begin{subfigure}[b]{0.25\textwidth}
        \centering
        \includegraphics[width=0.97\textwidth]{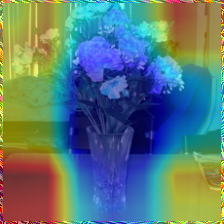}
        \caption{Predicted: maypole}
    \end{subfigure}

    \begin{subfigure}[b]{0.25\textwidth}
        \centering
        \includegraphics[width=0.97\textwidth]{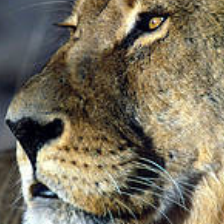}
        \caption{Ground-truth: lion}
    \end{subfigure}\hspace{0.025\textwidth}
    \begin{subfigure}[b]{0.25\textwidth}
        \centering
        \includegraphics[width=0.97\textwidth]{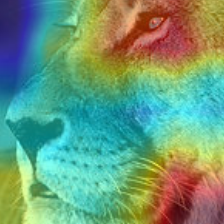}
        \caption{Predicted: lion}
    \end{subfigure}\hspace{0.025\textwidth}
    \begin{subfigure}[b]{0.25\textwidth}
        \centering
        \includegraphics[width=0.97\textwidth]{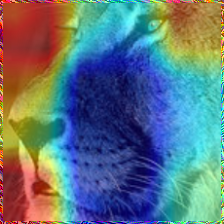}
        \caption{Predicted: maypole}
    \end{subfigure}
\renewcommand{\familydefault}{\rmdefault}
\caption{Grad-CAM for inputs from ImageNet. For each example, first the original image is shown, and then the visualizations for unattacked and attacked image. While the network correctly identifies key objects for classification in unattacked images, it concentrates on the image borders when given adversarial input.} \label{fig_saliency}
\end{figure*}

We computed such visualizations for the pretrained ResNet-50 from PyTorch Model Zoo, taking as input images from ImageNet. We consider both the cases with and without an adversarial framing. Few qualitative results are presented in Figure~\ref{fig_saliency}\footnote{We use the following Grad-CAM implementation:  \href{https://github.com/kazuto1011/grad-cam-pytorch}{\texttt{github.com/kazuto1011/grad-cam-pytorch}}.}.

\subsection{Classifier's input resizing}\label{subsec_resizing}

Our method does not modify pixels of the original input (with dimensions $h \times w$) and only adds a framing around it. This results in dimensions of the classifier input becoming $(h + 2W) \times (w + 2W)$ where $W$ is framing's width. This is fine for most of the state-of-the-art image classification architectures; however, to make sure the approach also works for classifiers with fixed input size, we conducted experiments with attacking the ImageNet classifier for several image resizing strategies:
\begin{enumerate}[label=(\alph*),leftmargin=2\parindent]
    \item \label{vanilla} no resizing, input dimensions are changed (\textbf{Vanilla}). The framing is trained with Algorithm~\ref{alg:adversarial_framing}.
    \item first the framing is added, and then the whole image is rescaled back to $h \times w$ (\textbf{Frame \& Resize, F\&R}). We use the same framing as in \ref{vanilla}.
    \item \label{rf} the image is first scaled to $(h - 2W) \times (h - 2W)$ and then the framing is added, so that size is again $h \times w$ (\textbf{Resize \& Frame, R\&F}). We train the framing separately because the number of parameters is smaller than in \ref{vanilla}.
    \item framing is put on the original image, occluding its border pixels; the size remains unchanged (\textbf{Occlude}). We use the same framing as in \ref{rf}.
\end{enumerate}

While we see differences in results, all the variants prove very efficient for $W = 4$. Compared to other resizing strategies, performance is especially degraded in \textbf{Frame \& Resize}. This is expected since the adversarial framing itself is resized and mixed with neighbouring pixels there.

Based on these results, if one can change input dimensions, \textbf{Vanilla} approach performs the best, and otherwise \textbf{Resize \& Frame} leads to the highest error rate. Results are shown in Table~\ref{imagenet_resizing}.

\section{Related work}
\subsection{Universal partially-affecting attacks}

Since existing attacks are quite different from our approach, it is hard to perform a direct comparison. However, we try to compare our work with universal partially-affecting attacks using localized patches.
% which we think are the closest to our approach.
We are aware of two works that perform these kind of attacks, LaVAN \cite{lavan} and Adversarial patch \cite{adversarial_patch}. Both methods were tested on ImageNet and hence we will focus on that case.

Unfortunately, each of these works consider different percentages of the image pixels that may be altered.
%As a result, we would like to recall some of our results.
We thus first recall our results for various framing sizes and then relate it to results from other works.
With \textbf{AF} of width 1, we use less than $2\%$ of the image's pixels and accuracy in untargeted setting drops to $10.53\%$. For $W = 2$, we use less than $3.5\%$ of the image's pixels and the accuracy is $0.44\%$ only. Finally, for $W = 4$ we use less than $7\%$ of the image's pixels to make the classifier almost completely confused ($0.1\%$ accuracy) in untargeted setting and achieve $99.66\%$ average success rate in targeted setting.

In LaVAN, a patch occluding about $2\%$ of the image is used (which is comparable to our \textbf{AF} of width 1). Their universal attack has success rate $74.1\%$ in targeted setting. When they use the same patch to measure untargeted performance, they change the output class of the classifier for only $78.9\%$ of data, which suggests that the accuracy of the classifier is higher than $10.53\%$ achieved by our method.
% When they use the same patch to measure untargeted performance, they change the output class of the classifier for only 78.9% of data, which suggests that the accuracy of the classifier is higher than $10.53\%$ achieved by our method.

Adversarial patch is a method that creates localized perturbations which can be deployed in a real world. The authors consider targeted setting only. They measure success rate as a function of percentage of pixels used. They need to occlude at least $10\%$ of pixels to obtain $90\%$ success rate.

%It is hard to reach a definitive conclusion here; however, it seems that our method performs on par or better than prior approaches.
As mentioned before, the comparison to prior works is burdensome due to the differences in shape, localization and design of other approaches. However, when we put all these characteristics aside and focus on the performance in respect to the ratio of perturbed pixels to the original ones, it seems that our method performs better than prior approaches. Additionally, our method is shown to generalize to videos.

\subsection{Attacking video classifiers}
Although extensive literature exists on attacks against image classifiers, we are aware of only a few works on video classifier attacks~\cite{wei2018sparse,video,rey2018targeted}.
While resulting in successful attacks, these approaches are fully-affecting and hence introduce adversarial artifacts in the video. In contrast, output from our attack contains the original video and no information is lost. Moreover, the framing is constant over all video frames, removing any ``flickering'' effect that could potentially be distracting to viewers.

\begin{table}[t]
\centering
% \small
\tabcolsep=0.14cm
\begin{tabular}{l|cccc}
 \toprule
 \textbf{Attack} & $W = 1$ & $W = 2$ & $W = 3$ & $W = 4$ \\
 \midrule
 \textbf{None} & \multicolumn{4}{c}{76.13\%} \\
 \midrule
 \textbf{Vanilla} & \textbf{10.53\%} & \textbf{0.44\%} & \textbf{0.11\%} & \textbf{0.1\%} \\
 \textbf{F\&R} & $56.12\%$ & $20\%$ & $5.32\%$ & $1.19\%$ \\
 \textbf{R\&F} & $33.87\%$ & $1.09\%$ & $0.15\%$ & \textbf{0.1\%} \\
 \textbf{Occlude} & $43.78\%$ & $3.2\%$ & $0.33\%$ & $0.12\%$ \\
 \bottomrule
\end{tabular}
\vskip 0.1in
\caption{Accuracies of ImageNet classifier attacked using adversarial framing with different resizing strategies.}
  \label{imagenet_resizing}
\end{table}

%%%%%%%%%%%%%%%%%%%%%%%%%%%%%%%%%%%%%%%%%%%%%%%%%%%%%%%%%%%%%%%%%%%%%%%%%%%%%%%%%%%%%%%%%%%%%%%%%%
% Conclusion
%%%%%%%%%%%%%%%%%%%%%%%%%%%%%%%%%%%%%%%%%%%%%%%%%%%%%%%%%%%%%%%%%%%%%%%%%%%%%%%%%%%%%%%%%%%%%%%%%%
\section{Conclusion}
In this work, we present a simple method for attacking both image and video classifiers. The proposed attack is universal (\textit{i.e.}\ the same adversarial framing can be applied in different images or videos), efficient and effective. Moreover, our method does not modify the original content of the input and only adds a small border to surround it.

\section{Acknowledgments}
Micha\l{} Zaj\k{a}c is co-financed by National Centre for Research and Development as a part of EU supported Smart Growth Operational Programme 2014-2020 (POIR.01.01.01-00-0392/17-00).
Konrad \.Zo\l{}na is financially supported by National Science Centre, Poland (\mbox{2017/27/N/ST6/00828}, \mbox{2018/28/T/ST6/00211}).

We modified the repository provided by \cite{zolna2019classifier} to implement our method (see \href{https://github.com/kondiz/casme}{\texttt{github.com/kondiz/casme}} for their code).

This is an extended version of the paper published at 33rd AAAI Conference on Artificial Intelligence (see \href{https://doi.org/10.1609/aaai.v33i01.330110077}{\texttt{doi.org/10.1609/aaai.v33i01.330110077}}).

%References and End of Paper
%These lines must be placed at the end of your paper

\bibliography{main}
\bibliographystyle{aaai}
\end{document}